\documentclass[a4paper, 10pt, conference]{ieeeconf}      
\IEEEoverridecommandlockouts
\usepackage{cite}
\usepackage{amsmath,amssymb,amsfonts}
\usepackage[linesnumbered,ruled,vlined]{algorithm2e}
\usepackage{graphicx}
\usepackage{textcomp}
\usepackage{xcolor}
\usepackage{soul}
\usepackage{caption}
\usepackage{subcaption}
\usepackage{balance}
\usepackage[hidelinks]{hyperref}
\usepackage{array}

\newcommand{\PreserveBackslash}[1]{\let\temp=\\#1\let\\=\temp}
\newcolumntype{C}[1]{>{\PreserveBackslash\centering}p{#1}}

\def\BibTeX{{\rm B\kern-.05em{\sc i\kern-.025em b}\kern-.08em
    T\kern-.1667em\lower.7ex\hbox{E}\kern-.125emX}}

\makeatletter
\newcommand{\linebreakand}{%
  \end{@IEEEauthorhalign}
  \hfill\mbox{}\par
  \mbox{}\hfill\begin{@IEEEauthorhalign}
}
\makeatother

\title{\LARGE \bf Self-Reconfigurable V-shape Formation \\of Multiple UAVs in Narrow Space Environments

}

\author{Duy Nam Bui$^{1}$, Manh Duong Phung$^{2}$, Hung Pham Duy$^{1}$
\thanks{$^{1}$Duy Nam Bui and Hung Pham Duy are with VNU University of Engineering and Technology, Hanoi, Vietnam. {e-mail: \tt\footnotesize duynam@ieee.org; hungpd@vnu.edu.vn}.}
\thanks{$^{2}$Manh Duong Phung is with Fulbright University Vietnam, Ho Chi Minh City, Vietnam. {e-mail: \tt\footnotesize duong.phung@fulbright.edu.vn}.}
}

\begin{document}

\maketitle

\begin{abstract}
This paper presents the design and implementation of a self-reconfigurable V-shape formation controller for multiple unmanned aerial vehicles (UAVs) navigating through narrow spaces in a dense obstacle environment. The selection of the V-shape formation is motivated by its maneuverability and visibility advantages. The main objective is to develop an effective formation control strategy that allows UAVs to autonomously adjust their positions to form the desired formation while navigating through obstacles. To achieve this, we propose a distributed behavior-based control algorithm that combines the behaviors designed for individual UAVs so that they together navigate the UAVs to their desired positions. The reconfiguration process is automatic, utilizing individual UAV sensing within the formation, allowing for dynamic adaptations such as opening/closing wings or merging into a straight line. Simulation results show that the self-reconfigurable V-shape formation offers adaptability and effectiveness for UAV formations in complex operational scenarios. 
\end{abstract}


\section{Introduction}
Unmanned aerial vehicles (UAVs) has gained significant attention in recent years due to their potential applications in various fields, including surveillance, search and rescue operations, and infrastructure inspection \cite{8682048,9990164}. One crucial aspect of UAV operations is their ability to navigate and maintain formations effectively, especially in complex environments with obstacles. The formation control of multiple UAVs thus plays a vital role in achieving coordination and efficient mission execution \cite{Anderson,9990236}.

\begin{figure}
    \centering
    \includegraphics[width=0.48\textwidth]{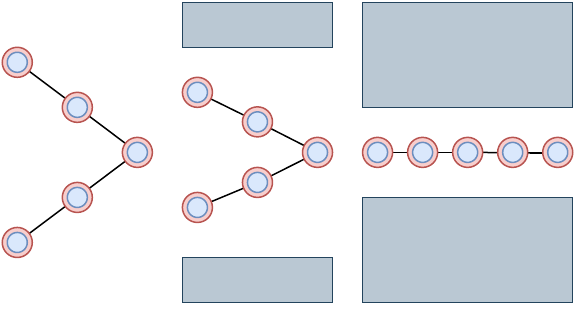}
    \caption{The self-reconfigurable V-shape formation can adjust its shape and navigate through narrow passages.}
    \label{fig:idea}
\end{figure}

Formation control in multi-robot systems refers to the coordination and control strategies used to form, maintain, and transform formations among a group of robots\cite{736776,Balch2000}. In scenarios with dense obstacles or narrow spaces, UAV formations need to adapt and reconfigure their shape to navigate through or around obstacles \cite{7487747,Dang2019,8843165}, as described in Figure \ref{fig:idea}. This self-reconfigurable capability enables the UAVs to overcome challenging terrain, narrow passages, or complex structures, thereby enhancing their maneuverability and overall mission success. One commonly adopted formation shape is the V-shape configuration, which offers advantages in terms of stability, visibility, and aerodynamic efficiency \cite{Dang2019,Mirzaeinia2019}. In \cite{Dang2019}, a formation control algorithm is proposed for multi-UAV systems where the UAVs autonomously adjust their positions within a V-shape to avoid collisions with obstacles. In \cite{8793765}, a splitting and merging algorithm is proposed for multi-robot formations in environments presented by static and dynamic obstacles. The work in \cite{8594438} presents a switching strategy of a region-based shape controller for a swarm of robots to deal with the obstacle-avoidance problem in complex environments. 

Recently, advancements in path planning and obstacle avoidance techniques have contributed to the development of self-reconfigurable formation control algorithms. In \cite{8843165}, the angle-encoded particle swarm optimization algorithm is developed for the formation of multiple UAVs used in vision-based inspection of infrastructure. By incorporating constraints related to flight safety and visual inspection, the path and formation can be combined to provide trajectory and velocity profiles for each UAV. The work in \cite{FENG2022} developed a novel optimization method for multi-UAV formation to achieve rapid and accurate reconfiguration under random attacks. In \cite{Gao2022}, multi-UAV reconfiguration problems are modeled as an optimal problem with task assignment and control optimization. However, the focus of their work was on maintaining a fixed formation shape rather than self-reconfiguration in the presence of obstacles.


In this work, we propose a self-reconfigurable V-shape formation control algorithm for multiple UAVs operating in narrow space where the formation cannot maintain its initial shape when moving through this space. The algorithm allows the UAVs to form, maintain and reconfigure the desired V-shape formation by expanding/shrinking its two V-wings to avoid collisions with obstacles and maintain distances among the UAVs. According to the proposed design of distributed behaviors, the V-shape formations can open/close wings or merge into a straight line. Based on it, the UAVs can navigate through narrow passages, bypass obstacles, and optimize their trajectory in accordance to environmental conditions. The main contributions of our work are twofold: (i) develop a self-reconfiguration strategy that can adjust its shape in narrow space environments; and (ii) propose reconfiguration behaviors that navigate UAVs and maintain their shape.

The remaining sections of this paper are structured as follows. Section \ref{sec:design} describes the design of the V-shape formation. Section \ref{sec:control} presents our implementation of the behavior-based controller. Section \ref{sec:result} shows simulation results. Our paper ends with conclusions drawn in Section \ref{sec:conclusion}.
\section{V-Shape Formation Design} \label{sec:design}
The V-shape formation is chosen due to its advantages in improving maneuverability and enhancing visibility for UAVs. Its modeling and design principles are presented as follows. 
\subsection{UAV model}
\begin{figure}
    \centering
    \includegraphics[width=0.25\textwidth]{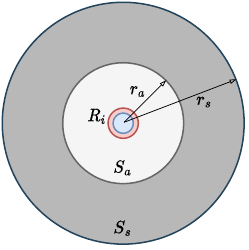}
    \caption{The sensing range $r_s$ and alert range $r_a$ ($r_a<r_s$) of a UAV $R_i$}
    \label{fig:model}
\end{figure}
 
The formation consists of $n$ identical UAVs, each equipped with sensory modules for positioning and navigation such as Lidar, GPS and inertial measurement unit (IMU). The UAV is also equipped with a communication module that allows peer-to-peer communication among the UAVs. At height $h$, a UAV $R_i$ is modeled as a particle moving in a 2D plane located at that height with position $p_i$ and heading angle $\psi_i$. The single-integrator kinematic model of UAV $R_i$ can be expressed as follows:
\begin{equation}
    \dot{p}_i = u_i,
\end{equation}
where $u_i=[u_{ix},u_{iy}]^T$ is the velocity vector of UAV $R_i$. The heading angle $\psi_i$ then can be obtained as:
\begin{equation}
    \psi_i=\text{atan2}(u_{iy},u_{ix}).
\end{equation}

The communication range of each UAV $R_i$ is divided into two areas including the sensing area $S_s$ with radius $r_s$ and the alert area $S_a$ with radius $r_a<r_s$ so that $S_a\subset S_s$, as illustrated in Figure \ref{fig:model}.

\subsection{V-shape formation}
\begin{figure}
    \centering
    \includegraphics[width=0.3\textwidth]{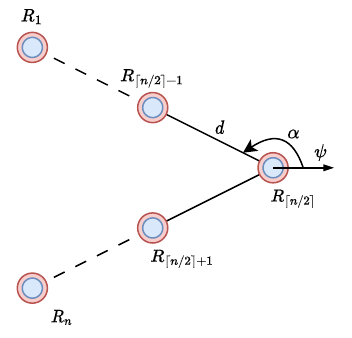}
    \caption{Illustration of the V-shape formation}
    \label{fig:vshape}
\end{figure}
In this work, the V-shape formation is constructed by two wings\cite{Dang2019}, as shown in Figure \ref{fig:vshape}. The wings are described by the desired distances between consecutive UAVs, $d$, and the bearing angle between the formation heading and each wing, $\alpha$. Without loss of generality, choose UAV $R_l$, with $l=\left\lceil{n}/{2}\right\rceil$, as the leader UAV located at the forefront of the formation. The desired distance $d_i$ and angle $\alpha_i$ between UAV $R_i$, $i\neq l$, and UAV $R_l$ are determined as follows:
\begin{equation}
\begin{aligned}
    d_i&=d\left\vert l-i\right\vert,\\
    \alpha_{i}&=\left\{ \begin{array}{cc}
\psi_{l}+\alpha & \text{if }i<l\\
\psi_{l}-\alpha & \text{if }i>l
\end{array}\right.
\end{aligned}
\label{eqn:desired}
\end{equation}

Thus, the desired position of UAV $R_i$ can be obtained as follows:
\begin{equation}
    p_i^d=p_l+d_i\left[\begin{array}{c}
\cos\alpha_{i}\\
\sin\alpha_{i}
\end{array}\right].
\label{eqn:desired_pose}
\end{equation}
\section{Distributed Formation Control Strategy} \label{sec:control}
The formation control strategy aims to form, maintain and self-reconfigurate the V-shape formation in response to obstacles and narrow passages during UAV navigation. This adaptation is achieved through either expanding/shrinking two wings of the V-shape. It allows the UAV formation to navigate safely within confined spaces without encountering collisions. The proposed algorithm operates based on the use of distributed behavior-based control and artificial potential field approaches so that individual UAVs can make decisions and adjust the formation shape as needed. The strategy consists of two parts: maintaining formation and reconfigurating formation.

\begin{figure*}
    \centering
    \begin{subfigure}[b]{0.32\textwidth}
    \includegraphics[width=\textwidth]{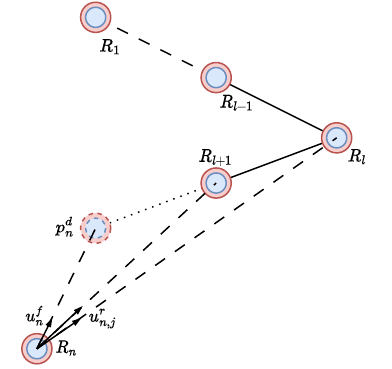}
    \caption{Disruption in formation}
    \label{fig:reconfig0}
    \end{subfigure}
    \begin{subfigure}[b]{0.32\textwidth}
    \includegraphics[width=\textwidth]{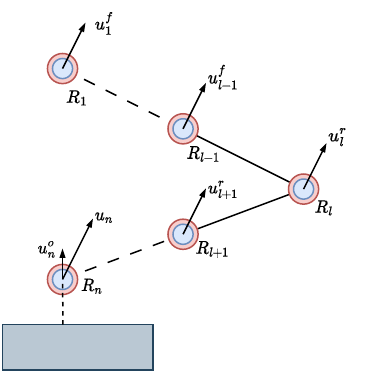}
    \caption{Obstacles from one side}
    \label{fig:reconfig1}
    \end{subfigure}
    \begin{subfigure}[b]{0.32\textwidth}
    \includegraphics[width=\textwidth]{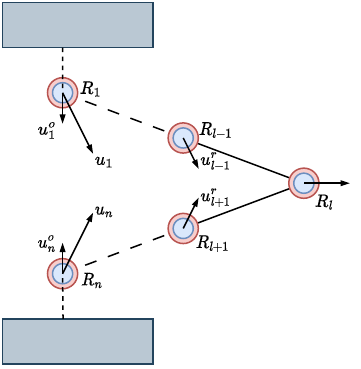}
    \caption{Obstacles from both sides}
    \label{fig:reconfig2}
    \end{subfigure}
    \caption{Self-reconfiguration of the V-shape formation based on the mechanism of pliers or scissors}   
    \label{fig:reconfiguration}
\end{figure*}

\subsection{Formation maintenance strategy}
Behavior-based control is the approach that combines a set of distributed control modules, called behaviors, to achieve the desired objective \cite{Mataric2008, 736776}. In this work, the UAV formation is maintained based on the combination of the following behaviors.
\subsubsection{Formation behavior}
The formation behavior aims to guide UAVs to achieve their desired positions within the predefined formation. According to \eqref{eqn:desired_pose}, the desired position $p_i^d$ of UAV $R_i$ in the formation can be obtained. Inspired by \cite{Dang2019,MirzaeeKahagh2020}, we define the formation behavior as follows:
\begin{equation}
    u_i^f=-k_f\left(p_i- p_i^d\right) + u_l,
    \label{eqn:uf}
\end{equation}
where $k_f>0$ is a positive formation gain. 

\subsubsection{Goal reaching behavior}
This behavior navigates the formation towards the desired location. To accomplish this objective, a target-tracking controller is formulated based on the relative position between the leader UAV and the goal. Let $p_g$ be the goal position that the formation needs to reach. The goal reaching behavior is constructed as follows: 
\begin{equation}
    u_i^g=-k_g\left(p_i - p_g\right),
    \label{eqn:ug}
\end{equation}
where $k_g>0$ is a positive tracking gain.

\subsubsection{Obstacle avoidance behavior}
During operation, the formation must avoid obstacles present in the environment. Let $p_{io_h}$ be the closest point on the boundary of obstacle $o_h$ within the sensing range of UAV $R_i$. When that UAV senses obstacle $o_h$, it will create a thrust to maneuver and avoid the obstacle. The thrust is directed as follows:
\begin{equation}
    u_{ih}^{o}=\left\{ \begin{array}{cc}
-k_o\left(\dfrac{1}{d_{io_h}^2} - \dfrac{1}{r_s^2}\right)\dfrac{p_i-p_{io_h}}{\left\Vert p_i-p_{io_h}\right\Vert}, & \text{if } d_{io_h} < r_s\\
0. & \text{otherwise}\\
\end{array}\right.
\end{equation}
where $k_o>0$ is a positive gain; $d_{io_h}$ is the distance between UAV $R_i$ and obstacle $o_h$. When considering all obstacles, the obstacle avoidance behavior of UAV $R_i$ are obtained as follows:
\begin{equation}
    u_i^o=\sum_{h=1}^m{u_{ih}},
    \label{eqn:uo}
\end{equation}
where $m$ is number of observable obstacles within the sensing range of UAV $R_i$.

\subsubsection{Collision avoidance behavior}
Apart from avoiding obstacles, the control algorithm also needs to adjust the UAV positions to avoid collision among them. To address this, we propose that UAVs $R_i$ and $R_j$ that are not in the same wing but within each other's sensing area, i.e., $\left\Vert p_{ij}\right\Vert < r_{s}$, will exert a repulsive force to preventing the UAVs from entering the alert area $S_a$. Let $p_{ij}=p_i-p_j$. The collision avoidance behavior is determined as follows:
\begin{equation}
    u_{ij}^{c}=k_{c}\dfrac{e^{-\beta_{c}\left(\left\Vert p_{ij}\right\Vert -r_{a}\right)}}{\left\Vert p_{ij}\right\Vert -r_{a}}\dfrac{p_i-p_j}{\left\Vert p_i-p_j\right\Vert},
    \label{eqn:uc}
\end{equation}
where $k_c>0$ is a positive collision gain.


\subsection{Self-reconfigurable formation strategy} 
\label{sec:reconfig}
\begin{figure*}
    \centering
    \begin{subfigure}[b]{\textwidth}
        \includegraphics[width=\textwidth]{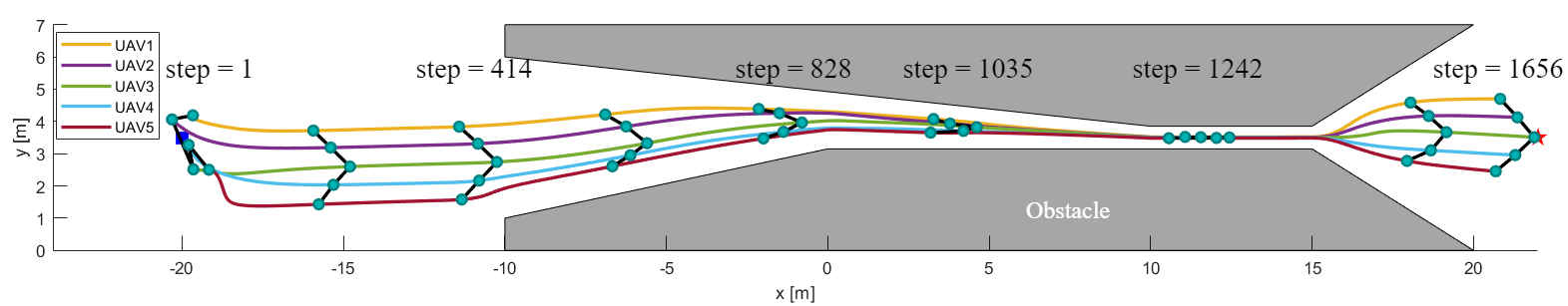}
        \caption{Trajectories of the UAVs in the formation}
        \label{fig:motion}
    \end{subfigure}
    \begin{subfigure}[b]{\textwidth}
        \includegraphics[width=\textwidth]{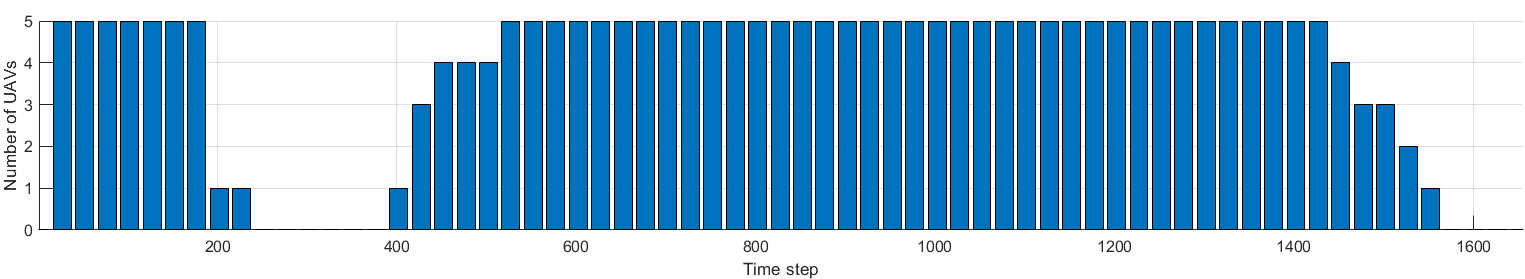}
        \caption{Number of UAVs activating reconfiguration behaviors over time}
        \label{fig:number}
    \end{subfigure}
    \caption{Simulation result of the V-shape formation moving through a narrow passage}
    \label{fig:result}
\end{figure*}

Inspired by the mechanics of pliers and scissors, which alter their shape through the application of opposing forces on their handle arms, the V-shape formation can open or close its wings based on the exertion of external forces. In our work, the force is produced based on the difference in the distances among the UAVs and their desired distances. Specifically, in case the formation encounters disruptions arising from improper positioning, as illustrated in Figure \ref{fig:reconfig0} where UAV $R_n$ deviates from the alignment, the combined force acts to guide it toward its desired location.

In the scenario depicted by Figure \ref{fig:reconfig1} where a force is exerted from one side, UAV $R_n$ responds by generating a thrust $u_n^o$ to avoid a potential collision with obstacles thus resulting in the control signal $u_n$. Accordingly, other UAVs on the same V-wing, including the leader UAV, also adjust their position based on the behavior control signal $u_i^r$. As the leader UAV $R_l$ changes its position, the UAVs on the opposing wing realign themselves by formation behavior $u_i^f$. As a result, the whole UAV formation tends to move towards the other side.

In another scenario, when obstacles impact the formation from both opposing sides, as depicted in Figure \ref{fig:reconfig2}, they affect UAVs $R_1$ and $R_n$, leading to the generation of obstacle avoidance behaviors denoted as $u_1^o$ and $u_n^o$. Other UAVs on the same wing respond by generating reconfiguration behaviors $u_i^r$ that adjust the UAVs' position accordingly. As a result, the V-shape formation is able to shrink its wing to travel through narrow passages.

In our work, the aforementioned reconfiguration idea is implemented by the following equation:

\begin{equation}
    u_{ij}^{r}=k_{r}\dfrac{\left|\left\Vert p_{ij}\right\Vert -d_{ij}\right|^{\beta_r}}{\left(\left\Vert p_{ij}\right\Vert -r_{a}\right)^{2}}\dfrac{p_i-p_j}{\left\Vert p_i-p_j\right\Vert},
    \label{eqn:ur}
\end{equation}
where $k_r>0$ is a positive reconfiguration gain, $\beta_r>0$ is the smoothness factor, $d_{ij}$ is the desired distance between two UAVs $R_i$ and $R_j$ n the same wing, $d_{ij}=d\left\vert i-j\right\vert$. In (\ref{eqn:ur}), the term $\left|\left\Vert p_{ij}\right\Vert -d_{ij}\right|$ enables the UAVs to adjust their positions so that the desired distances among the UAVs are maintained. This behavior is also used as a collision avoidance behavior for the UAVs in the same wing.



\subsection{Overall strategy}


The overall distributed control strategy is obtained by combining the behaviors from all UAVs as follows:
\begin{equation}
    u_{i}=\left\{ \begin{array}{cc}
u_{i}^{g}+u_{i}^{r}+u_{i}^{c}+u_{i}^{o}, & \text{if assigned as leader}\\
u_{i}^{f}+u_{i}^{r}+u_{i}^{c}+u_{i}^{o}. & \text{otherwise}
\end{array}\right.
\end{equation}

According to this function, the behaviors are automatically triggered in response to external influences or disturbances encountered by the formation. Once the desired state is achieved, the behavioral values are maintained resulting in stable operation of the UAV formation.
\section{Results and Discussion} \label{sec:result}
In this section, we evaluate the performance of the proposed control strategy through different simulation scenarios. 
The source code of the strategy can be found at {\fontfamily{pcr}\selectfont \url{https://github.com/duynamrcv/reconfigurable_vshape}}.

\begin{figure*}
    \centering
    \begin{subfigure}[b]{0.49\textwidth}
    \includegraphics[width=\textwidth]{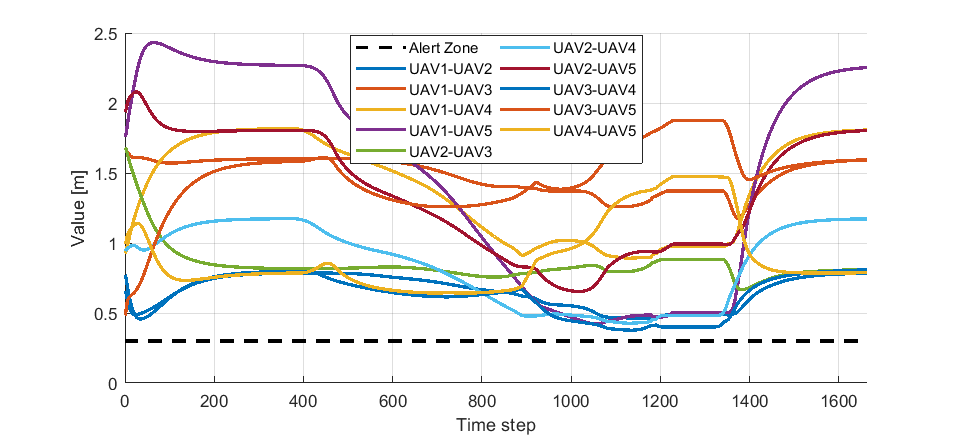}
    \caption{Distances between each pair of UAVs over time}
    \label{fig:distance}
    \end{subfigure}
    \begin{subfigure}[b]{0.49\textwidth}
    \includegraphics[width=\textwidth]{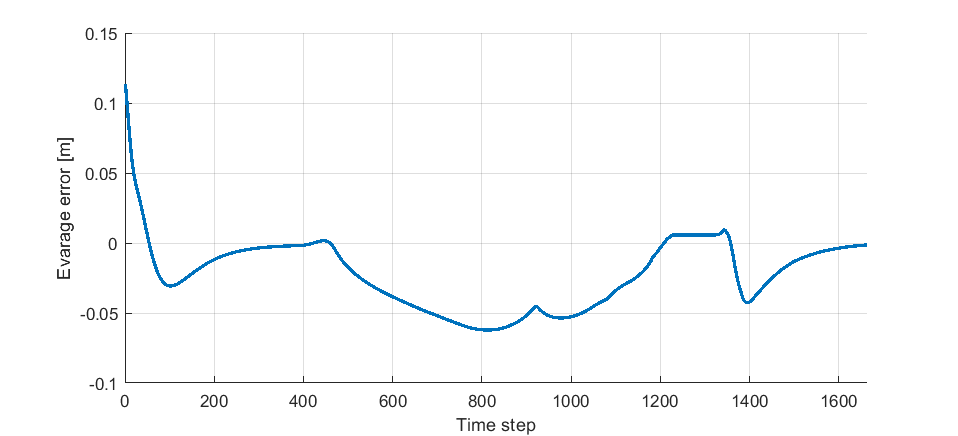}
    \caption{The average distance error of UAV formation}
    \label{fig:error}
    \end{subfigure}
    \begin{subfigure}[b]{0.49\textwidth}
    \includegraphics[width=\textwidth]{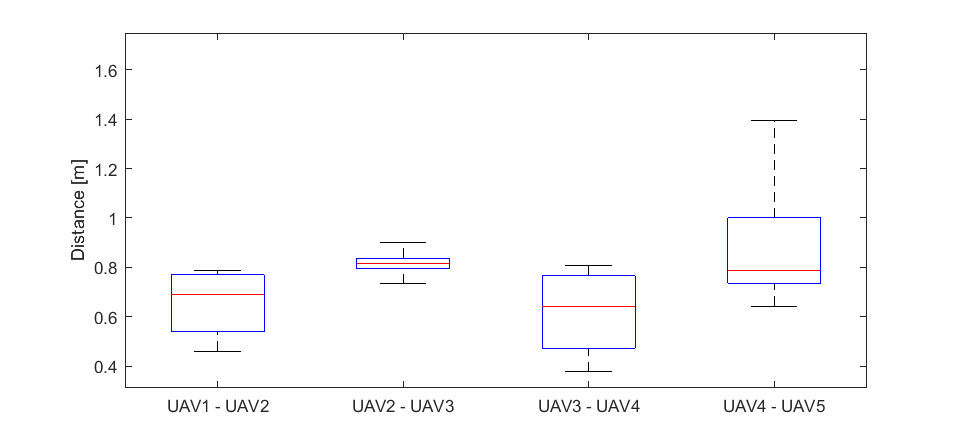}
    \caption{The distance between consecutive UAVs}
    \label{fig:mean}
    \end{subfigure}
    \begin{subfigure}[b]{0.49\textwidth}
    \includegraphics[width=\textwidth]{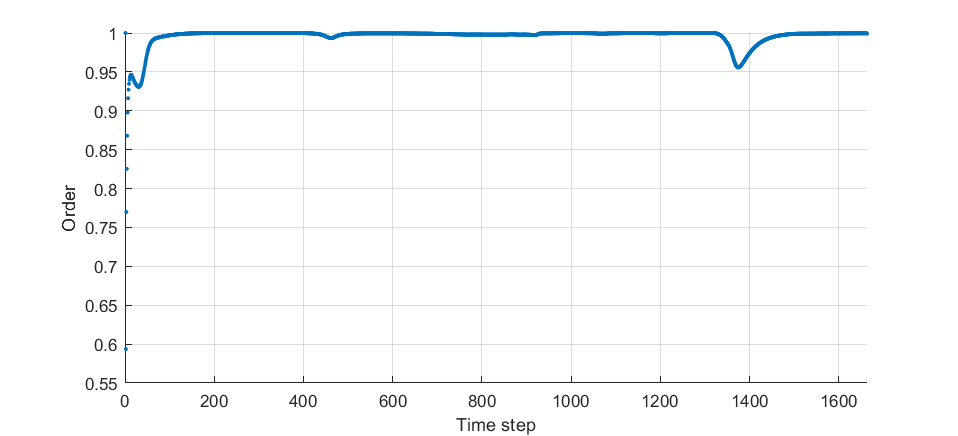}
    \caption{Values of the \textit{order} metric $\Phi$ over time}
    \label{fig:heading}
    \end{subfigure}
    \caption{Evaluation of the proposed algorithm}
    \label{fig:eval}
\end{figure*}

\begin{table*}
\centering
\caption{Statistical evaluation of the proposed strategy for several different scenarios}
\label{tbl:sta}
\begin{tabular}{C{1.2cm}C{1.5cm}C{1cm}C{1cm}C{2cm}C{1.8cm}C{3.2cm}}
\hline
Scenario & Number UAVs & $d$ & $\alpha$    & Average error (m) & Min distance (m) & Average distance of consecutive UAVs (m) \\ \hline
1     & 3        & 1.0   & $3\pi/4$             & 0.12333   & 0.48557   & 0.98788                      \\
2     & 5        & 1.0   & $3\pi/4$    & 0.12068   & 0.37383   & 0.96575                      \\
3     & 3        & 0.8   & $5\pi/6$    & 0.10942   & 0.48988   & 0.87599                      \\
4     & 3        & 1.0   & $4\pi/5$    & 0.13666   & 0.47581   & 1.0943                       \\
5     & 5        & 0.8   & $3\pi/4$    & 0.111     & 0.41279   & 0.88913     \\ \hline                
\end{tabular}
\end{table*}

\subsection{Simulation setup}
In the simulation, the UAV has the alert radius $r_a = 0.3$ m and the sensing radius $r_s = 2$ m. The control period is set at $0.02$ s. The maximum speed of each UAV is $2.0 \text{ m/s}$. The V-shape formation is defined with $d=0.8 \text{ m}$ and $\alpha=0.3\pi/4 \text{ rad}$. In our evaluation, 5 UAVs with V-shape formation are operated in the area of 46 m $\times$ 7 m with two large obstacles arranged to form a narrow passage as shown in Figure \ref{fig:result}.

\subsection{Results}

Figure \ref{fig:result} shows the trajectories of the UAVs moving in the environment under the guidance of the proposed distributed controller. Initially, the UAVs are randomly positioned around a starting point (step 1). They then self-adjust to form the desired V-shape formation (step 414) based on the control signals generated by the formation and reconfiguration behaviors. As they encounter obstacles, the UAVs start to adjust their formation. This includes deforming the formation (steps 414-1035) and transitioning to a straight-line formation (step 1242) to navigate through narrow gaps. Upon successfully circumventing the obstacles, the UAVs readjust to form the desired V-shape and proceed toward the goal position (step 1656). The result can be verified in the simulation video shown in footnote \footnote{Simulation video: {\fontfamily{qcr}\selectfont
\url{https://youtu.be/_6u7yMNOySc}}}. 

Figure \ref{fig:number} shows the number of UAVs activating their reconfiguration behavior over time. It can be seen that the UAVs activate this behavior when shaping the formation at the initialization stage, and during the process of adjusting their formation to adapt to the environment structure.

The statistical results of the evaluation of the proposed strategy are depicted in Figure \ref{fig:eval}. Figure \ref{fig:distance} shows the distances between the UAVs over time. It can be seen that those distances are all greater than the alert radius, which confirms the effectiveness of the control algorithm in avoiding collision among the UAVs.

Figure \ref{fig:error} presents the average distance error of the UAV formation over time. Initially, the error is large since the UAVs have not formed the desired shape. After the control algorithm realigns the UAV to their desired positions, the error quickly converges toward zero. While the formation navigates through the narrow passage, the error remains small, less than $0.06$ m. Figure \ref{fig:mean} shows the average distance between consecutive UAVs. It can be seen that the average distance fluctuates around the desired value for the V-shaped formation ($d=0.8$ m), which is desirable for the formation.

To further evaluate the performance of the proposed controller, an \textit{order} metric $\Phi$ that measures the similarity in the UAVs' direction is used \cite{Vicsek1995}. It takes the values in range $[0, 1]$ and is computed as follows:
\begin{equation}
    \Phi=\dfrac{1}{n}\left\Vert\sum_{i=1}^n{\left[\cos\psi_i, \sin\psi_i\right]^T}\right\Vert.
    \label{eq:order}
\end{equation}

According to (\ref{eq:order}), the order metric $\Phi$ is close to 1 when all UAVs have the same heading angle. Figure \ref{fig:heading} shows the value of $\Phi$ in our simulation. It can be seen that $\Phi$ is close to 1 during the movement of the formation, even when the formation avoids obstacles or traverses through a narrow passage. Changes in the heading angle increase when exiting the passage since the UAVs in the formation need to realign to the origin shape. However, the order metric then quickly converges to 1 when the UAVs form their desired formation. The results thus confirm the validity of the proposed control algorithm.

To further evaluate the performance of the proposed method, simulations on various scenarios such as narrow passages of varying widths and dense obstacle areas have been conducted. In addition, the number of UAVs and V-shape parameters, the desired distance, $d$, and the desired bearing angle, $\alpha$, are also varied. The results, including the average formation error, the closest distance between two UAVs, and the average distance between consecutive UAVs, are summarized in Table {\ref{tbl:sta}}. It is evident that the average formation error approximates $0.1m$, which is sufficient for the formation to maneuver in narrow spaces. Furthermore, the minimum distance between any pair of UAVs is larger than the collision threshold, $r_a$, and thus meets the requirement for collision avoidance. Finally, the average distance between consecutive UAVs closely aligns with the desired value, indicating the stability in the formation shape. These results confirm the validity and effectiveness of our proposed control strategy.
\section{Conclusion} \label{sec:conclusion}
In this work, we have presented a new behavior-based controller to address the problem of UAV formation in narrow space environments. We developed several behaviors for each UAV and then proposed a function to combine them. Our approach allows the UAVs to form a V-shape formation with the capability to adjust their wings to avoid obstacles and travel through narrow passages. A number of simulation evaluations have been conducted and the results show that our control strategy is not only able to navigate the UAVs to form the desired V-shape formation but also provide them with the capability to reconfigure themselves to circumvent obstacles, avoid collisions, and traverse narrow passages in complex environments.
\section*{Acknowledgement}
Duy Nam Bui was funded by the Master, PhD Scholarship Programme of Vingroup Innovation Foundation (VINIF), code VINIF.2022.Ths.057.
\balance

\bibliographystyle{ieeetr}  
\bibliography{ref}

\begin{thebibliography}{10}

\bibitem{8682048}
H.~Shakhatreh, A.~H. Sawalmeh, A.~Al-Fuqaha, Z.~Dou, E.~Almaita, I.~Khalil,
  N.~S. Othman, A.~Khreishah, and M.~Guizani, ``Unmanned aerial vehicles
  ({UAVs}): A survey on civil applications and key research challenges,'' {\em
  IEEE Access}, vol.~7, pp.~48572--48634, 2019.

\bibitem{9990164}
N.~Duong Thi~Thuy, D.~Nam~Bui, M.~Duong~Phung, and H.~Pham~Duy, ``Deployment of
  {UAVs} for optimal multihop ad-hoc networks using particle swarm optimization
  and behavior-based control,'' in {\em 2022 11th International Conference on
  Control, Automation and Information Sciences (ICCAIS)}, pp.~304--309, 2022.

\bibitem{Anderson}
B.~D.~O. Anderson, B.~Fidan, C.~Yu, and D.~Walle, ``{UAV} formation control:
  Theory and application,'' in {\em Lecture Notes in Control and Information
  Sciences}, pp.~15--33, Springer London, 2008.

\bibitem{9990236}
H.~P. Quang, T.~Nguyen~Dam, V.~N. Hoang, and H.~Pham~Duy, ``Multi-{UAV}
  coverage strategy with v-shaped formation for patrol and surveillance,'' in
  {\em 2022 11th International Conference on Control, Automation and
  Information Sciences (ICCAIS)}, pp.~487--492, 2022.

\bibitem{736776}
T.~Balch and R.~Arkin, ``Behavior-based formation control for multirobot
  teams,'' {\em IEEE Transactions on Robotics and Automation}, vol.~14, no.~6,
  pp.~926--939, 1998.

\bibitem{Balch2000}
T.~Balch, ``Hierarchic social entropy: An information theoretic measure of
  robot group diversity,'' {\em Autonomous Robots}, vol.~8, no.~3,
  pp.~209--238, 2000.

\bibitem{7487747}
J.~Alonso-Mora, E.~Montijano, M.~Schwager, and D.~Rus, ``Distributed
  multi-robot formation control among obstacles: A geometric and optimization
  approach with consensus,'' in {\em 2016 IEEE International Conference on
  Robotics and Automation (ICRA)}, pp.~5356--5363, 2016.

\bibitem{Dang2019}
A.~D. Dang, H.~M. La, T.~Nguyen, and J.~Horn, ``Formation control for
  autonomous robots with collision and obstacle avoidance using a rotational
  and repulsive force{\textendash}based approach,'' {\em International Journal
  of Advanced Robotic Systems}, vol.~16, p.~172988141984789, May 2019.

\bibitem{8843165}
V.~Hoang, M.~Phung, T.~Dinh, Q.~Zhu, and Q.~Ha, ``Reconfigurable multi-{UAV}
  formation using angle-encoded pso,'' in {\em 2019 IEEE 15th International
  Conference on Automation Science and Engineering (CASE)}, pp.~1670--1675,
  2019.

\bibitem{Mirzaeinia2019}
A.~Mirzaeinia, M.~Hassanalian, K.~Lee, and M.~Mirzaeinia, ``Energy conservation
  of v-shaped swarming fixed-wing drones through position reconfiguration,''
  {\em Aerospace Science and Technology}, vol.~94, p.~105398, Nov. 2019.

\bibitem{8793765}
H.~Zhu, J.~Juhl, L.~Ferranti, and J.~Alonso-Mora, ``Distributed multi-robot
  formation splitting and merging in dynamic environments,'' in {\em 2019
  International Conference on Robotics and Automation (ICRA)}, pp.~9080--9086,
  2019.

\bibitem{8594438}
D.~Roy, A.~Chowdhury, M.~Maitra, and S.~Bhattacharya, ``Multi-robot virtual
  structure switching and formation changing strategy in an unknown occluded
  environment,'' in {\em 2018 IEEE/RSJ International Conference on Intelligent
  Robots and Systems (IROS)}, pp.~4854--4861, 2018.

\bibitem{FENG2022}
Q.~Feng, X.~Hai, B.~Sun, Y.~Ren, Z.~Wang, D.~Yang, Y.~Hu, and R.~Feng,
  ``Resilience optimization for multi-{UAV} formation reconfiguration via
  enhanced pigeon-inspired optimization,'' {\em Chinese Journal of
  Aeronautics}, vol.~35, pp.~110--123, Jan. 2022.

\bibitem{Gao2022}
C.~Gao, J.~Ma, T.~Li, and Y.~Shen, ``Hybrid swarm intelligent algorithm for
  multi-{UAV} formation reconfiguration,'' {\em Complex {\&} Intelligent
  Systems}, vol.~9, pp.~1929--1962, Oct. 2022.

\bibitem{Mataric2008}
M.~J. Matari{\'{c}} and F.~Michaud, {\em Behavior-Based Systems}, pp.~891--909.
\newblock Berlin, Heidelberg: Springer Berlin Heidelberg, 2008.

\bibitem{MirzaeeKahagh2020}
A.~M. Kahagh, F.~Pazooki, and S.~E. Haghighi, ``Obstacle avoidance in v-shape
  formation flight of multiple fixed-wing {UAVs} using variable repulsive
  circles,'' {\em The Aeronautical Journal}, vol.~124, pp.~1979--2000, Oct.
  2020.

\bibitem{Vicsek1995}
T.~Vicsek, A.~Czir{\'{o}}k, E.~Ben-Jacob, I.~Cohen, and O.~Shochet, ``Novel
  type of phase transition in a system of self-driven particles,'' {\em
  Physical Review Letters}, vol.~75, pp.~1226--1229, Aug. 1995.

\end{thebibliography}
\end{document}